\documentclass[letterpaper]{article} 
\usepackage{aaai23}
\usepackage{times}  
\usepackage{helvet}  
\usepackage{courier}  
\usepackage[hyphens]{url}  
\usepackage{graphicx} 
\usepackage{natbib}  
\usepackage{caption} 
\usepackage{algorithmicx}
\usepackage{algpseudocode}
\usepackage{algorithm}
\usepackage{amsmath}
\usepackage{amsthm}
\usepackage{amssymb}
\newtheorem{example}{Example}

\usepackage{multirow}
\usepackage{booktabs}
\usepackage{xcolor}

\usepackage{newfloat}
\usepackage{listings}
\DeclareCaptionStyle{ruled}{labelfont=normalfont,labelsep=colon,strut=off} 
\floatstyle{ruled}
\newfloat{listing}{tb}{lst}{}
\floatname{listing}{Listing}
\title{Outlier Explanation via Sum-Product Networks}
\author{
   Stefan Lüdtke\textsuperscript{\rm 1}, Christian Bartelt\textsuperscript{\rm 1} and Heiner Stuckenschmidt\textsuperscript{\rm 2}
}
\affiliations{
    \textsuperscript{\rm 1}Institute for Enterprise Systems, University of Mannheim, Germany\\
        \textsuperscript{\rm 2}Data and Web Science Group, University of Mannheim, Germany
}

\begin{document}

\maketitle

\begin{abstract}
Outlier explanation is the task of identifying a set of features that distinguish a sample from normal data, which is important for downstream (human) decision-making. 
Existing methods are based on beam search in the space of feature subsets. They quickly becomes computationally expensive, as they require to run an outlier detection algorithm from scratch for each feature subset.

To alleviate this problem, we propose a novel outlier explanation algorithm based on Sum-Product Networks (SPNs), a class of probabilistic circuits. Our approach leverages the tractability of marginal inference in SPNs to  compute outlier scores in feature subsets. 
By using SPNs, it becomes feasible to perform backwards elimination instead of the usual forward beam search, which is less susceptible to missing relevant features in an explanation, especially when the number of features is large. 
We empirically show that our approach achieves state-of-the-art results for outlier explanation, outperforming recent search-based as well as deep learning-based explanation methods. 
\end{abstract}
\section{Introduction}

The identification of uncommon or anomalous samples (outliers) in a dataset is an important task in data science. Outliers can, for example, indicate defective products in  quality control \cite{stojanovic2016big}, intrusions in networks \cite{garcia2009anomaly}, or potential medical conditions in health records \cite{carrera2019online}. 
Many outlier detection methods have been proposed, including classical methods based on notions of distance or density \cite{liu2008isolation,scholkopf2001estimating} as well as  deep learning-based methods (see \citealt{pang2021deep} for a review).

A less well investigated, but natural question is that of \emph{outlier explanation}: Given a sample classified as an outlier, which properties of the sample are the cause for this classification, i.e., which properties are specifically anomalous?
This task has also been called \emph{outlying aspect mining}
 \cite{duan2015mining,vinh2016discovering,samariya2020new} or \emph{outlier interpretation} \cite{xu2021beyond,liu2018contextual}.

\begin{figure}[tb]
\centering
\includegraphics[scale=0.27]{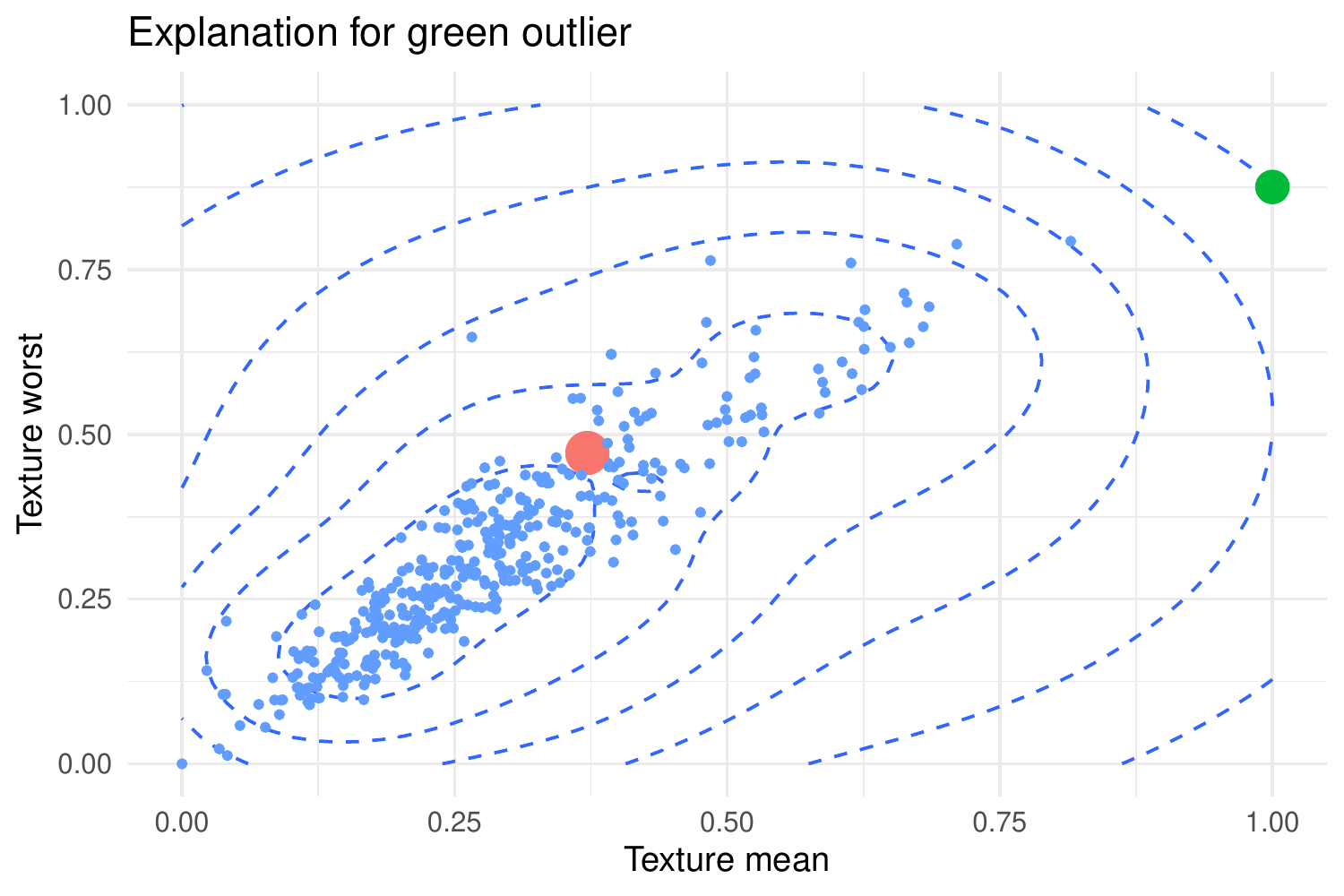}
\includegraphics[scale=0.27]{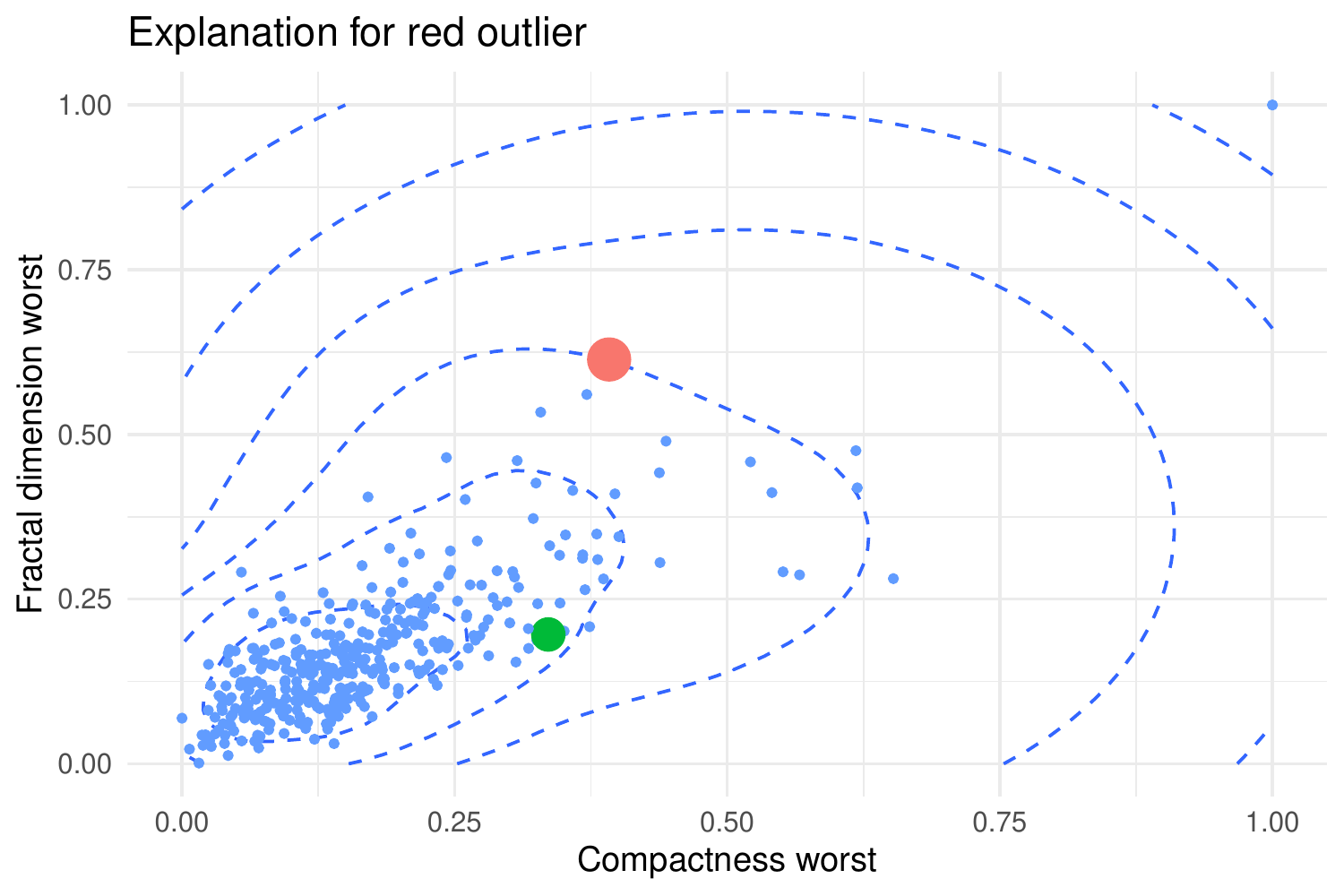}
\caption{Two explanations generated for outliers in the Wisconsin Breast Cancer dataset. Contours illustrate the (marginal) density of the SPN. Left: Features that best explain outlyingness of the green sample. Right: Feature that best explain outlyingness of the red sample. 
 The explanations indicate that the green sample has unusual texture, while the red sample has unusual shape.}
\label{fig:example}
\end{figure}

\begin{example}
\normalfont
The Wisconsin Breast Cancer dataset\footnote{Available at \url{odds.cs.stonybrook.edu/wbc}.} contains properties of cell nuclei of fine needle aspirates of breast mass. Outliers correspond to malignant cases. 
A support system for a physician should not only \emph{detect} potential outliers (malignant cases), but  also explain why it considers a sample an outlier, to increase the physician's confidence in the system and support her diagnosis and therapy decision.
Figure \ref{fig:example} shows two examples of such explanations. 
We can see that the green sample has unusual texture and the red sample has unusual shape, which can be relevant for downstream tasks like therapy decisions.
\end{example}

Most existing methods for outlier explanation are based on beam search to identify feature subsets in which the outlier score of a sample is maximal \cite{duan2015mining,vinh2016discovering,wells2019new,samariya2020new}. They require to run an outlier detection algorithm for each feature subset that is visited during beam search, which can be computationally very costly.

The contribution of this paper is a novel algorithm for outlier explanation based on \emph{Sum-Product Networks} (SPNs) \cite{poon2011sum}. SPNs are probabilistic models in which marginal inference is tractable. In this model, identification of feature subsets with high outlier score is fast, compared to existing methods: An SPN only needs to be trained once, and marginal probabilities in the SPN correspond to outlier scores in the respective feature subsets.
Furthermore, we propose to use a  backward selection search strategy instead of the usual forward beam search to identify feature subsets.
The option to use backward selection is enabled by the use of SPNs as underlying model: As runtime of outlier score computation in SPNs does not depend on the feature subset size,  it become possible to start the search for an explanation with a large feature subset and prune it iteratively.
We find that backward elimination is less susceptible to miss relevant features than beam search, especially for dataset with larger dimensionality, overall leading to more accurate explanations for high-dimensional data.

We extensively evaluate our approach on a number of synthetic and real-world outlier  explanation tasks. Our approach achieves state-of-the-art results for outlier explanation, outperforming recent search-based as well as deep learning-based explanation methods, while being computationally feasible.  

\section{Preliminaries and Related Work}

\subsection{Explainable Outlier Detection}
\label{subsec:outlier-detection}

\paragraph{Outlier Detection}
Outlier detection is the following unsupervised learning task: Given a dataset $\{\mathbf{x^{(1)}},\dots,\mathbf{x^{(n}}\}$, classify each sample as either normal or outlier.
Here, the samples $\mathbf{x} \in \mathcal{X}$ can be multivariate, e.g. $\mathcal{X} = \mathbb{R}^n$, but we consider the case where some or all dimensions are categorical as well. 
Outlier detection algorithms usually compute a \emph{scoring function} $f: \mathcal{X} \rightarrow \mathbb{R}$, which can be used for outlier classification by classifying all samples with $f(\mathbf{x}) > t$ as outliers, for a fixed threshold $t$. 
Classical methods include, for example, isolation forests \cite{liu2008isolation}, local outlier factor \cite{breunig2000lof} or one-class support vector machines \cite{scholkopf2001estimating}. 
More recently, deep neural networks have been used for this task \cite{pang2021deep}.

In this paper, we focus on probabilistic outlier detection, by assuming that \emph{normal} data was generated from a distribution $p(\mathbf{x};\theta)$ with parameters $\theta$. 
An outlier is a sample which is unlikely to be drawn from $p(\mathbf{x};\theta)$. 
That is, we use the scoring function $f(\mathbf{x}) = -p(\mathbf{x};\theta)$.
Parametric as well as non-parametric density estimators have been considered for $p(\mathbf{x};\theta)$, e.g., Gaussian mixtures \cite{pimentel2014review} or Kernel Density Estimation \cite{schubert2014generalized}. 
The parameters $\theta$ can be estimated from purely normal training data, in which case the task is also called \emph{novelty detection} \cite{pimentel2014review}. For the outlier detection task considered here, where the available data consists of normal as well as anomalous data, it is still customary to use the complete data to estimate $\theta$, assuming that outliers are rare and sparse so that they will still be assigned a low probability.

\paragraph{Outlier Explanation}
\emph{Outlier explanation} is the task of retrieving a subset of features in which the sample is specifically anomalous \cite{zhang2004hos,duan2015mining,wells2019new,samariya2020new}. 
More formally, let $D \subseteq \{1,\dots,n\}$ be a set of indices, and let $\mathbf{x}_D$ denote the  projection of $\mathbf{x}$ onto the subspace indicated by $D$. 
Outlier explanation is the task of identifying $D$, such that $f(\mathbf{x}_D)$ is maximized for a given sample $\mathbf{x}$. 

The naive approach of computing $f(\mathbf{x}_D)$ individually for each subspace $D \subseteq \{1,\dots,n\}$ quickly becomes infeasible due to the combinatorial explosion in $|D|$.
Therefore, existing methods \cite{zhang2004hos,duan2015mining,wells2019new,samariya2020new}  for outlier explanation usually perform a greedy beam search that iteratively adds dimensions to $D$. 
For example, \citet{duan2015mining} use a kernel density estimator (KDE) to compute outlier scores for each visited subspace. \citet{wells2019new} build on this work, replacing the KDE with a faster, grid-based density estimator. 
Still, these methods are computationally expensive as they need to run a density estimator individually for each investigated subspace \cite{samariya2020comprehensive}. 

To select the best explanation, simply returning the subspace $D$ where $f(\mathbf{x}_D$ is minimized is not usually not appropriate, because scoring functions for different dimensionalities are usually not directly comparable \cite{vinh2016discovering}. For example, the distance between samples generally increases when the number of dimensions increases, favoring high-dimensional subspaces as explanations.
Thus, \emph{dimensionality-unbiased} scores like \emph{z-score} normalization 
\begin{equation}
z(\mathbf{x},D,\mathbf{X}) = \frac{f(\mathbf{x}_D) - \mu(\mathbf{X}_D)}{\sigma(\mathbf{X}_D)}
\label{eq:zscore}
\end{equation}
w.r.t. the training dataset $\mathbf{X}$ or a rank transformation have been proposed \cite{duan2015mining}.
They allow to compare scores between different dimensionalities, thus allowing to identify a subspace $D$ in which $\mathbf{x}_D$ is most outlying, relative to other samples.
However, they are computationally expensive, as outlier scores need to be computed for all samples $\mathbf{x}' \in \mathbf{X}$, instead of only the query sample $\mathbf{x}$.

In contrast to search-based methods, explanations can also be obtained via algorithm-agnostic, local explainability methods from the area of supervised learning, e.g. LIME \cite{ribeiro2016should} or SHAP \cite{lundberg2017unified}.
They explain a prediction by assigning an importance value to each feature. 
Feature importance-based explanation methods specifically tailored towards the outlier detection task have been proposed as well \cite{liu2018contextual,xu2021beyond}. 
For example, ATON \cite{xu2021beyond} 
consists of an embedding layer and a subsequent self-attention layer, where attention weights represent contribution of each embedding dimension to the outlyingness of an outlier. From these weights, feature importance weights (in the original feature space) can be computed.
This method has been shown to produce more accurate explanations than general explainability methods (LIME, SHAP) for the task of outlier explanation. 

\subsection{Sum-Product Networks}

\paragraph{Representation}
A Sum-Product Network (SPN) \cite{poon2011sum} is a rooted directed acyclic graph representing a probability distribution over a sequence of random variables (RVs) $\mathbf{X} = X_1,\dots,X_n$.
Each node represents a distribution $p_N$ over a subset $\mathbf{X}_{\phi(N)} \subseteq \mathbf{X}$, where $\phi(N) \subseteq \{1,\dots,n\}$ is called the scope of the node $N$. In the following, $\text{ch}(N)$ denotes the children of node $N$. 
An SPN contains tree types of nodes: Leaf nodes, product nodes and sum nodes. A product node represents a factorized distribution $p(\mathbf{X}_{\phi(N)}) = \prod_{C \in \text{ch}(N)} p_C(\mathbf{X}_{\phi(C)})$.
A sum node represents a mixture distribution $p_N(\mathbf{X}_{\phi(N)}) = \sum_{C \in \text{ch}(N)} w_C \, p_C(\mathbf{X}_{\phi(C)})$. Finally, a leaf node directly represents a (tractable) univariate or multivariate distribution.
\emph{Decomposability} (children of product nodes have pairwise disjoint scopes) and \emph{completeness} (children of sum nodes have identical scope) ensure that an SPN actually represents a valid probability distribution. 
By definition, the distribution represented by an SPN is the distribution defined by its root node. 
 
Early research on SPNs focused on categorical distributions \cite{poon2011sum} or simple parametric leaf distributions, like Gaussians \cite{dennis2012learning}.   
More recently, SPNs with piecewise polynomial leaf distributions have been used to model continuous and mixed data \cite{molina2018mixed}.

\paragraph{Inference}
The appealing property of SPNs is that any marginal distribution $p(\mathbf{X'}{=}\mathbf{x'})$ for a subset $\mathbf{X'} \subset \mathbf{X}$ can be computed efficiently. Intuitively, this is possible because summation over the marginalized RVs can be ``pushed down'' into the leaf nodes of the SPN \cite{peharz2015theoretical}. 
Thus, marginal inference reduces to marginalization of the leaves and evaluating the internal nodes of the SPN once. As leaves are usually chosen such that marginal inference in leaf distributions is possible in constant time, marginal inference is linear in the number of nodes of the SPN. Specifically, when the leaf distributions are univariate, the value
of marginalized leaves can simply be set to 1.

\paragraph{Learning}
A number of different learning algorithms for SPNs have been proposed. 
Early learning algorithms focused on structure learning \cite{gens2013learning,vergari2015simplifying,peharz2013greedy,molina2018mixed}. Most prominently, LearnSPN \cite{gens2013learning} is a greedy structure learning algorithm, which creates a tree-structured SPN in a top-down fashion. 
It  recursively tests for independence of RVs (in which case it creates a product node and recurses), and otherwise clusters the data into subsets, creates a corresponding sum node and recurses. 
\citet{molina2018mixed} proposed an extension of LearnSPN which also works for continuous and mixed domains.
Recently, \citet{peharz2020random} proposed a learning algorithm which first initializes a random SPN structure and then learns parameters via EM. This way, parameter learning can leverage fast, parallel GPU computations, as shown by \citet{peharz2020einsum}.

\section{Explainable Outlier Detection via SPNs}
In this section, we present a novel  outlier detection model, and show how outlier explanations can be extracted from this model in a straightforward way.

\subsection{SPNs for Outlier Detection}

Probabilistic outlier detection methods use a scoring function of the form $f(\mathbf{x}) = -p(\textbf{x};\theta)$.
The main idea of this paper is to use an SPN to represent the joint density  $p(\textbf{x};\theta)$. This approach has several advantages, compared to existing outlier detection approaches:
\begin{itemize}
\item SPNs are powerful and expressive density estimators, reaching state-of-the-art performance in several density estimation tasks. Thus, they should be able to accurately learn $p(\textbf{x};\theta)$, leading to good density estimation (and subsequently, outlier detection) performance. 
\item SPNs can seamlessly handle mixed discrete-continuous domains, which is difficult for distance-based methods as well as methods using parametric distributions.
\item SPNs directly lend themselves to outlier explanation due to their tractable inference.  In fact, in this paper, we use them solely for this purpose.
\end{itemize}

\subsection{SPNs for Outlier Explanation}
\label{subsec:spn-explanation}

\begin{figure*}[tb]
\centering
\includegraphics[scale=0.35]{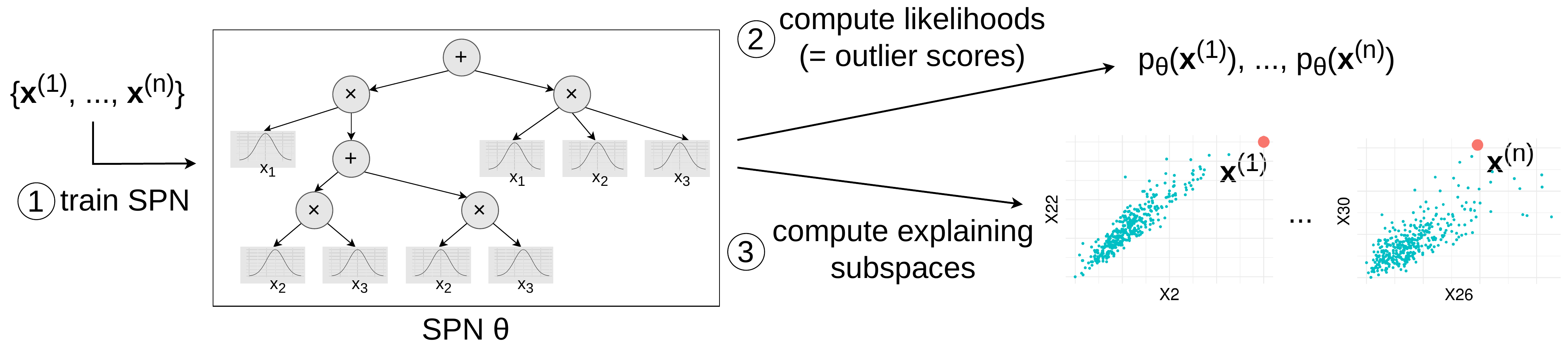}
\caption{Overview of proposed method for outlier detection and explanation. An SPN only needs to be trained once (1) and can subsequently be used to compute outlier scores (2), as well as anomalous feature subsets (i.e., explanations) via marginal inference in the SPN (3).}
\label{fig:abstract}
\end{figure*}

As discussed above, the central challenge in outlier explanation is to efficiently compute $f(\mathbf{x}_D)$ for subspaces $D$. For probabilistic outlier detection methods, this task is equivalent to computing a marginal distribution $f(\mathbf{x}_D) = p(\mathbf{x}_D; \theta)$. Such a marginal is obtained by integrating over all RVs $\mathbf{X} \setminus \mathbf{X}_D$. 
More formally, let $\bar{D} = \{1,\dots,n\}\setminus D$, and denote $\bar{D} = \{\bar{D}_1,\dots,\bar{D}_k\}$. The outlier score in subspace $D$ is given by
\begin{equation}
\begin{split}
& f(\mathbf{x}_D)  = p(\mathbf{x}_D;\theta)  \\
= & \int_{x_{\bar{D}_1}} \dots \int_{x_{\bar{D}_k}} p(\mathbf{x};\theta) \, \text{d}x_{\bar{D}_1} \dots \text{d}x_{\bar{D}_k}
\end{split}
\label{eq:marginal}
\end{equation}
Explicitly computing such marginals is intractable for many expressive density estimators. Instead, the strategy taken by existing outlier explanation methods \cite{duan2015mining,wells2019new} is to project the training samples to the subspace $D$, and estimate the parameters of the model $p(\mathbf{x}_D;\theta)$ from those samples. This approach is computationally expensive, as outlier scores are computed for many subspaces during beam search for the subspace in which the sample is most outlying. 

In SPNs, however, marginal inference is tractable: The time complexity of evaluating a marginal probability in Equation \ref{eq:marginal} is linear in the number of nodes of the SPN \cite{poon2011sum}, independently of the number of RVs that are marginalized---and irrespective of the number of original training samples, in contrast to approaches that perform parameter estimation for each subspace.

Hence, we propose the outlier detection and explanation model shown in Figure \ref{fig:abstract}: A single SPN representing the joint $p(\mathbf{x};\theta)$ is trained once, which can then subsequently be used to compute outlier scores as well as outlier explanations. 
Specifically, outlier explanations are computed via search in the feature subspaces, to identify the subspace where the outlier score of a sample is maximal.
In the following, we discuss the search strategy as well as the strategy for selecting the dimensionality of the explanation in more detail.

\paragraph{Search Strategies}
As computing outlier scores for all $2^n-1$ feature subspaces quickly becomes infeasible with increasing number of features $n$, a search strategy that only explores promising subspaces is required.
\textbf{Forward beam search}, which greedily adds features to the explanation has been used for this task before \cite{vinh2016discovering}. 
More specifically, the beam search keeps a set of $B$ hypotheses (feature subspaces). In each step and for each hypothesis, it greedily adds that feature to the hypothesis that maximizes the outlier score of the sample in the extended feature set. Search is carried out until a maximum depth $S$. The search algorithm is shown in Algorithm \ref{alg:beam-search}.

\begin{algorithm}[t]
\caption{forwardBeamSearch($\mathbf{x}$,$S$,$B$,$\theta$)}
\label{alg:beam-search}
\begin{algorithmic}
\State \textbf{Input:} Outlier $\mathbf{x}$ of dimensionality $n$, maximum explanation size $S$, beam width $B$, distribution parameters $\theta$ (e.g., as an SPN)
\State \textbf{Output:} For each $k \in \{1,\dots,n\}$, a subspace $D$ of size $k$ in which $x_D$ is most anomalous
\State{$D_1 \leftarrow \textsc{argLowestDensities}(B,D,\theta,\mathbf{x})$}  \Comment{Store $B$ most outlying dimensions}
\State{$D_1^{(\text{best})} \leftarrow \textsc{argLowestDensities}(1,D,\theta,\mathbf{x})$} \Comment{Overall most outlying dimension, needed as return value later}
\For{$k \in \{2,\dots,S\}$}
\State{$D_k \leftarrow \{\}$}
\For{$D_{k-1}^{(i)} \in D_{k-1}$}  \Comment{For all hypotheses, get all candidate subspaces of size $k$}
\State{$D_k \leftarrow D_k \cup \{D_{k-1}^{(i)} \cup d \,|\, d \in D\}$}
\EndFor
\State{$D_k \leftarrow \textsc{argLowestDensities}(B,D_k,\theta,\mathbf{x})$} \Comment{Keep only $B$ most outlying subspaces as hypotheses for next iteration}
\State{$D_k^{(\text{best})} \leftarrow \textsc{argLowestDensities}(1,D_k,\theta,\mathbf{x})$} \Comment{Store the most outlying subspace of size $k$ for returning it later}
\EndFor
\State \Return $D_1^{(\text{best})},\dots,D_S^{(\text{best})}$ 
\Function{argLowestDensities}{$B,D_k,\theta,\mathbf{x}$} \Comment{Get the $B$ subspaces from $D_k$ where $\mathbf{x}$ is least likely}
\State{$L \leftarrow \{ p(\mathbf{x}_d;\theta) \,|\, d \in D_k\}$}
\State \Return $\{d \,|\, d \in D_k, \text{rank}(p(\mathbf{x}_d;\theta),L) \leq B\}$
\EndFunction
\end{algorithmic}
\end{algorithm}

At depth $k$, each hypothesis consists of $k$ features, and $n-k$ features need to be explored (where $n$ is the overall number of features). Thus, up to depth $k$, $\sum_{i=1}^k n-k < n \, k$ feature subspaces are explored per hypothesis. In SPNs, computing an outlier score (a marginal probability density) amounts to evaluating the SPN (with $N$ nodes) once,  resulting in an overall time complexity of beam search-based explanation of $\mathcal{O}(N\, n\, S)$, where $S$ is the maximum search depth.

Intuitively, beam search works well when a sample that has high outlier score in a feature set of size $k$ also has a high outlier score in one of the subsets of size $k-1$. When this is not the case, beam search can fail to find reasonable explanations, as pointed out by \citet{xu2021beyond}.

To alleviate this problem, we propose a top-down, \textbf{backward elimination} search strategy to identify explanatory subsets. Instead of greedily adding dimensions, the search algorithm starts with the full feature set, and then greedily removes one feature at a time, so that in resulting feature subspace, the outlyingness of the sample is maximal (compared to all other subspaces of that size). The algorithm is shown in Algorithm \ref{alg:explain-backward}.
Intuitively, when a sample is an outlier in $k$-dimensional subspace, it cannot be a complete inlier in any $(k+1)$-dimensional subspace. Thus, starting from high dimensionality and only removing features can lead to more accurate results than bottom-up beam search.

At iteration $k$ of backward elimination, the feature subset consists of $n-k$ features. For each of the $n-k$ subsets of size $n-k-1$, an outlier score needs to be computed. The algorithm runs for $n$ iterations, resulting in $\sum_{k=0}^n (n-k) < n^2$ explored subsets.   Thus, overall runtime complexity of backward elimination is $\mathcal{O}(n^2\, N)$, where $N$ is the number of nodes of the SPN.


\begin{algorithm}[t]
\caption{backwardElimination($\mathbf{x}$,$\theta$)}
\label{alg:explain-backward}
\begin{algorithmic}
\State \textbf{Input:} Outlier $\mathbf{x}$ of dimensionality $n$, distribution parameters $\theta$ (e.g., as an SPN)
\State \textbf{Output:} For each $k \in \{1,\dots,n\}$, a subspace $D$ of size $k$ in which $x_D$ is most anomalous
\State{$D_n \leftarrow \{1,\dots,n\}$}
\For{$k \in \{n-1,\dots,1\}$}
\State{$d_k \leftarrow \underset{d \in D}{\text{argmax }} p(\mathbf{x}_{D_k \setminus d};\theta)$}
\State{$D_{k-1} \leftarrow D_k \setminus \{d_k\}$}
\EndFor
\State \Return $D_1,\dots,D_{n-1}$ 
\end{algorithmic}
\end{algorithm}

\paragraph{Dimensionality Selection}
Both beam search and backward elimination result in an outlier score for each visited feature subset. As a last step, one of the subsets needs to be selected as explanation. Simply selecting the subset with lowest outlier score might not be optimal, because scores for different dimensionalities are usually not directly comparable. Specifically, the densities $p(\mathbf{x}_D;\theta)$ will typically be smaller for larger dimensionality of $\mathbf{x}_D$. 
 \citet{vinh2016discovering} introduce \emph{dimensionality-unbiasedness} as a desideratum for outlier scores to allow for such comparison. Dimensionality-unbiasedness can be achieved, for example, by z-score transformation of the score in each subspace $D$, w.r.t.\ the scores of all samples in $D$ (see Equation \ref{eq:zscore}). However, these transformations are computationally inefficient as outlier scores need to be computed for all samples instead of only the query sample.

\begin{figure}[t]
\centering
\includegraphics[scale=0.5]{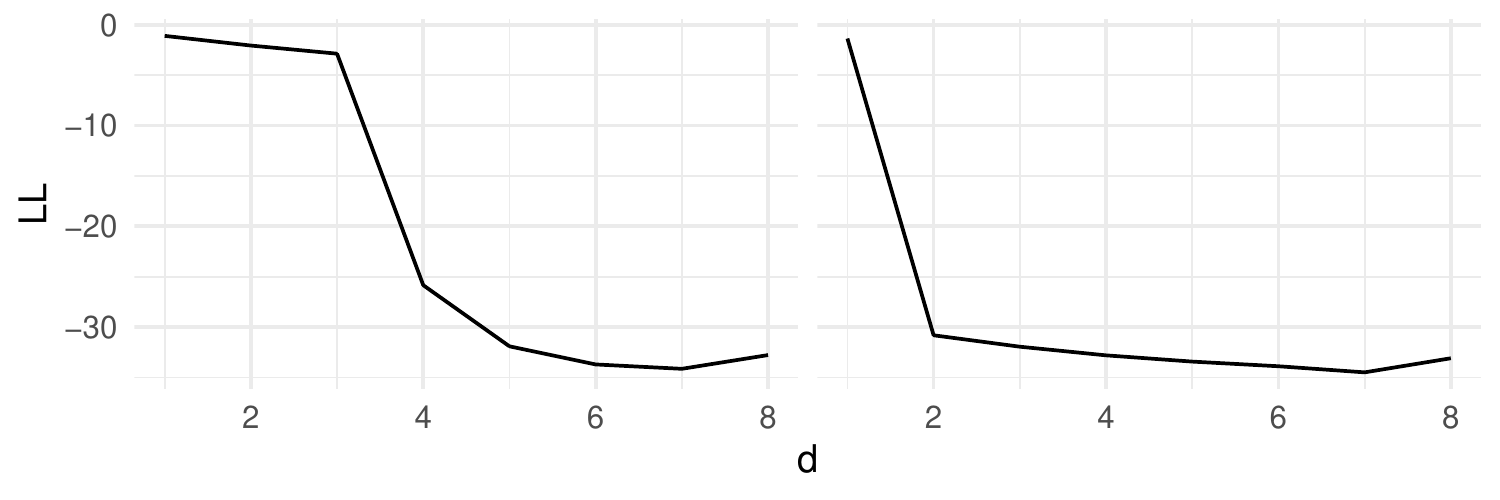}
\caption{Two examples of minimal log probability density of a sample, for different feature subsets sizes $d$ in the WBC dataset. In the left example, the true explanatory subspace has $d=4$, and $d=2$ in the right example. This can be easily identified by the drop in LL.}
\label{fig:steep-ll}
\end{figure}

Instead, we propose to use the \emph{elbow} method to select the optimal feature subset size (which has been, for example, used for determining the optimal number of clusters in k-means clustering \cite{aggarwal2015data}):
In real datasets, we often observe a large difference between the  minimal log density of all examined feature subsets of size $k$ and $k+1$ for a given sample, as shown in Figure \ref{fig:steep-ll}. In this case, we assume the subspace of size $k+1$ to be the explanation for that sample.
More concretely, we compute differences between subsequent lowest log densty, and then return the lowest-dimensional subspace where the difference is larger than a threshold $\kappa$. When a difference of at least $\kappa$ never occurs, we return the single feature with lowest univariate density. 

\section{Experimental Evaluation}
\label{sec:exp-evaluation}

Goal of the experiments was to evaluate the outlier explanation performance of our proposed method. Specifically, we aimed at answering the following research questions:
\begin{itemize}
\item[\textbf{Q1}] How accurate are the  explanations provided by our SPN-based approach for synthetic and real-world datasets, compared to state-of-the-art methods?
\item[\textbf{Q2}] How do the forward beam search and backward elimination search strategies for SPN-based outlier explanation compare, w.r.t.\ explanation performance?
\item[\textbf{Q3}] How does the proposed ebow method for dimensionality selection compare with the previously used method based on z-score transformation, w.r.t.\ explanation performance?
\item[\textbf{Q4}] How does runtime of our SPN-based approach scale w.r.t.\ data dimensionality, compared to state-of-the-art methods?
\end{itemize}

\subsection{Data Sets}

In our experiments, we used two groups of datasets: 

\paragraph{Synthetic Outlier Explanation Datasets} Evaluating outlier \emph{explanations} is not straightforward due to the lack of ground truth explanations. Therefore, we first used 21 synthetic datasets\footnote{Available at \url{www.ipd.kit.edu/mitarbeiter/muellere/HiCS}.} created by \citet{keller2012hics} for the purpose of evaluating outier ranking algorithms.  Each dataset consists of  10, 20, 30, 40, 50, 75 or 100 features (3 datasets per number of features) and contains 1000 samples, 19 to 136 of which are outliers. The datasets were created in such a way that each outlier is easily detectable  in a pre-defined, 2- to 5-dimensional  feature subset (which varies between outliers), but is an inlier in any lower-dimensional projection of the data. Goal of outlier explanation is to retrieve exactly those feature indices for each outlier. 
\paragraph{Real-World Outlier Explanation Datasets} Additionally, we evaluated outlier explanation performance on nine real-world datasets\footnote{Available at \url{github.com/xuhongzuo/outlier-interpretation}.} provided by \citet{xu2021beyond}. 
 To cope with the lack of ground-truth explanations, they created explanation labels for a set of real-world datasets as follows: First, each dataset was reduced to its ten first principal components. Then, for each dataset and each feature subset of that dataset, three  outlier detection algorithms (Isolation Forests \cite{liu2008isolation}, COPOD \cite{li2020copod} and HBOS \cite{goldstein2012histogram}) were applied to the subspace. The explanation label of an outlier was defined to be the feature subset where the outlier score is maximal (w.r.t. the algorithm). 
As a result of this procedure, each dataset has three distinct explanation labels per outlier, corresponding to the three outlier detection algorithms.  
From the available twelve datasets, we selected those nine datasets where at least one of the three outlier detection algorithms could achieve more than 0.5 ROC AUC, to ensure that the notion of outliers (and thus outlier explanations) is sensible.

\subsection{Experiments}

We compared our SPN-based outlier explanation algorithm to the following state-of-the-art outlier explanation algorithms:
\begin{itemize}
\item \textbf{ATON} \cite{xu2021beyond}, a state-of-the-art neural network model for outlier explanation based on attention.
\item \textbf{COIN} \cite{liu2018contextual} is an outlier explanation method which  fits a set of classifiers that separate outliers from clusters of nearby normal data, and uses the weights in the classifiers as feature importance values. 
\item \textbf{SiNNE} \cite{samariya2020new} is the latest contribution in a line of search-based outlier detection algorithms including \cite{vinh2016discovering} and \cite{wells2019new}. Instead of its predecessors, the approach uses a \emph{dimensionality-unbiased} outlier score function that does not require post-hoc normalization. 
\end{itemize}

We used implementations of these algorithms provided by \citet{xu2021beyond} \footnote{github.com/xuhongzuo/outlier-interpretation}.
Our SPN-based outlier detection algorithm was implemented in Python. We used the SPFlow library \cite{Molina2019SPFlow} for fitting and inference in SPNs. We used the LearnSPN algorithm \cite{gens2013learning} for SPN structure learning.

All SPN learning hyperparameters were set to fixed values across all experiments and datasets as follows:
We used Gaussian leaf distributions for real features and categorical leaf distributions for categorical features. 
During row splits, the data was partitioned via Expectation Maximization for Gaussian Mixture Models, using 2 mixture components. The Randomized Dependence Coefficient (RDC) \cite{lopez2013randomized} was used as independence test, setting $\alpha = 0.6$. The threshold for fitting leaves was set to $m=200$ samples to prevent overfitting. For beam search, we used a fixed beam width of 10, and set the threshold to $\kappa = \text{exp}(1)$.

This choice of SPN hyperparameters was based on \citet{vergari2015simplifying}, who found these hyperparameters to perform well on a set of 20 benchmark datasets (which are different from the datasets investigated here). Optimization of these hyperparameters on a validation set is possible and could improve SPN performance further. However, hyperparameter tuning was not attempted here as these fixed parameters already achieved good performance.

\section{Results}

\subsection{Outlier Explanation Performance}

\setlength{\tabcolsep}{5pt}

\begin{table}[t]
\begin{small}
\centering
\begin{tabular}{rrr|rrr}
  \toprule
D & SPN-fw & SPN-bw & ATON & COIN & SiNNE \\ 
  \midrule
10 & 0.867 (2) & 0.799 (5) & 0.806 (4) & \textbf{0.933 (1)} & 0.86 (3) \\ 
  20 & \textbf{0.668 (1)} & 0.646 (4) & 0.589 (5) & 0.667 (2) & 0.65 (3) \\ 
  30 & 0.562 (2) & \textbf{0.676 (1)} & 0.497 (4) & 0.427 (5) & 0.54 (3) \\ 
  40 & 0.399 (2) & \textbf{0.634 (1)} & 0.348 (3) & 0.261 (4) & - \\ 
  50 & 0.351 (2) & \textbf{0.682 (1)} & 0.3 (3) & 0.227 (4) & - \\ 
  75 & 0.355 (2) & \textbf{0.698 (1)} & 0.205 (3) & 0.158 (4) & - \\ 
  100 & 0.267 (2) & \textbf{0.611 (1)} & 0.154 (3) & 0.118 (4) & - \\ 
  \midrule
  Mean & 0.496 (2) & \textbf{0.678 (1)} & 0.414 (3) & 0.399 (4) & - \\  
   \bottomrule
\end{tabular}
\end{small}
\caption{Outlier explanation performance (F1 score of retrieved relevant dimensions and F1 score rank) for synthetic datasets. SPN-fw and SPN-bw denote SPN-based outlier explanation with forward and backward search strategies, respectively. SiNNE did not finish in less than 5,000 seconds for $D \geq 30$. }
\label{tbl:synthetic-results}
\end{table}

\paragraph*{Synthetic Data}
To assess \textbf{Q1} and \textbf{Q2}, we first evaluated the quality of the explanations (in terms of F1 score of retrieved features) on the synthetic datasets. We evaluated both forward beam search and backward elimination search. 

Table \ref{tbl:synthetic-results} shows F1 scores of the different outlier explanation methods. For each data dimensionality $D$, mean F1 scores of the three datasets of that dimensionality are reported. 
Regarding \textbf{Q1}, both SPN-based approaches outperformed the state-of-the-art methods (except for $D=10$), with an increasingly large difference in F1 for increasing $D$. SPN-bw (SPN with backward elimination search) is the only method where F1 score did not decrease substantially for larger data dimensionality, achieving good explanation performance even for $D=100$. 
With regards to the two search strategies (\textbf{Q2}), it can be seen that backward elimination outperformed beam search for higher-dimensional cases. We suspect that this is due to the fact that beam search is susceptible to missing relevant dimensions when their number increases (and the beam width stays constant), whereas backward elimination is more stable w.r.t. dimensionality.

\begin{table}[t!]
\centering
\begin{scriptsize}
\begin{tabular}{llllll}
  \toprule
dataset & SPN-fw & SPN-bw & ATON & COIN & SiNNE \\ 
  \midrule
\multirow{3}{*}{arrhythmia} & \textbf{0.742 (1)} & 0.726 (2) & 0.676 (3) & 0.367 (5) & 0.564 (4) \\ 
   & \textbf{0.635 (1)} & 0.577 (3) & 0.596 (2) & 0.398 (5) & 0.499 (4) \\ 
   & 0.695 (2) & \textbf{0.751 (1)} & 0.557 (3) & 0.273 (5) & 0.473 (4) \\ 
   \midrule
  \multirow{3}{*}{ionosphere} & \textbf{0.644 (1)} & 0.488 (4) & 0.622 (3) & 0.629 (2) & 0.482 (5) \\ 
   & 0.59 (2) & 0.452 (5) & \textbf{0.671 (1)} & 0.573 (3) & 0.454 (4) \\ 
   & \textbf{0.658 (1)} & 0.564 (4) & 0.618 (3) & 0.647 (2) & 0.433 (5) \\ 
   \midrule
  \multirow{3}{*}{letter} & \textbf{0.701 (1)} & 0.519 (5) & 0.665 (3) & 0.562 (4) & 0.668 (2) \\ 
   & 0.641 (2) & 0.388 (5) & \textbf{0.664 (1)} & 0.554 (4) & 0.614 (3) \\ 
   & \textbf{0.778 (1)} & 0.752 (2) & 0.545 (4) & 0.403 (5) & 0.616 (3) \\ 
   \midrule
  \multirow{3}{*}{optdigits} & \textbf{0.754 (1)} & 0.45 (5) & 0.671 (2) & 0.607 (4) & 0.654 (3) \\ 
   & \textbf{0.725 (1)} & 0.472 (5) & 0.672 (2) & 0.593 (4) & 0.622 (3) \\ 
   & \textbf{0.887 (1)} & 0.871 (2) & 0.557 (4) & 0.298 (5) & 0.58 (3) \\ 
   \midrule
  \multirow{3}{*}{pima} & 0.589 (2) & 0.538 (5) & \textbf{0.673 (1)} & 0.553 (4) & 0.588 (3) \\ 
   & 0.632 (2) & 0.515 (5) & \textbf{0.65 (1)} & 0.586 (3) & 0.557 (4) \\ 
   & \textbf{0.747 (1)} & 0.656 (2) & 0.531 (3) & 0.415 (5) & 0.441 (4) \\ 
   \midrule
  \multirow{3}{*}{satimage} & 0.604 (2) & \textbf{0.612 (1)} & 0.585 (3) & 0.429 (4) & 0.429 (5) \\ 
   & 0.661 (2) & 0.59 (3) & \textbf{0.664 (1)} & 0.539 (4) & 0.41 (5) \\ 
  & 0.746 (2) & \textbf{0.823 (1)} & 0.541 (3) & 0.247 (5) & 0.442 (4) \\ 
  \midrule
  \multirow{3}{*}{wbc} & \textbf{0.718 (1)} & 0.63 (2) & 0.604 (3) & 0.56 (5) & 0.57 (4) \\ 
   & 0.552 (2) & 0.447 (5) & \textbf{0.601 (1)} & 0.461 (4) & 0.499 (3) \\ 
   & \textbf{0.679 (1)} & 0.659 (2) & 0.579 (4) & 0.639 (3) & 0.502 (5) \\ 
   \midrule
  \multirow{3}{*}{wineRed} & 0.436 (3) & 0.366 (5) & \textbf{0.661 (1)} & 0.429 (4) & 0.505 (2) \\ 
   & 0.432 (4) & 0.367 (5) & \textbf{0.652 (1)} & 0.45 (3) & 0.493 (2) \\ 
   & \textbf{0.491 (1)} & 0.407 (4) & 0.481 (2) & 0.408 (3) & 0.361 (5) \\ 
   \midrule
  \multirow{3}{*}{wineWhite} & 0.526 (3) & 0.454 (4) & \textbf{0.619 (1)} & 0.436 (5) & 0.531 (2) \\ 
   & 0.469 (4) & 0.428 (5) & \textbf{0.605 (1)} & 0.497 (3) & 0.528 (2) \\ 
 & \textbf{0.569 (1)} & 0.529 (2) & 0.479 (3) & 0.38 (5) & 0.388 (4) \\ 
  \midrule
  Mean & \textbf{0.641 (1)} & 0.557 (3) & 0.609 (2) & 0.479 (5) & 0.515 (4) \\
   \bottomrule
\end{tabular}
\end{scriptsize}
\caption{Outlier explanation performance (F1 score of retrieved relevant dimensions and F1 score rank) for real-world datasets. The three rows for each dataset correspond to the three ground truth explanation labels. SPN-fw and SPN-bw denote SPN-based outlier explanation with forward and backward search strategies, respectively. }
\label{tbl:xu-results}
\end{table}

\paragraph*{Real Data}

Next, we evaluated outlier explanation performance on the real-world datasets processed by \citet{xu2021beyond}.
The results for ATON, COIN and SiNNE were taken directly from the paper introducing ATON \cite{xu2021beyond}. 
Table \ref{tbl:xu-results} shows the empirical results. For these datasets, our SPN-based approach (with forward search) outperformed the state-of-the-art in 17 out of 27 cases (63 \%).

With respect to \textbf{Q2}, forward beam search generally outerformed backward elimination for these datasets, which is consistent with results for the synthetic data: 
Keep in mind that the data were preprocessed by  \citet{xu2021beyond} such that they were at most 10-dimensional.
For such low-dimensional data, forward beam search (with a beam width of 10) was still able to identify explanations correctly.

Overall, the empirical results are encouraging: For the high-dimensional (synthetic) data, our SPN-based approach achieved a new state-of-the-art, and for the (low-dimensional) real-world data, our approach still outperformed state-of-the-art methods in 63\% of the cases.

\subsection{Dimensionality Selection}

\begin{table}[t]
\centering
\begin{small}
\begin{tabular}{rrrrrrr}
  \toprule
& \multicolumn{3}{l}{Forward Beam Search} & \multicolumn{3}{l}{Backward Elimination}\\
 \cmidrule(lr){2-4}  \cmidrule(lr){5-7} 
$D$ & elbow-1 & elbow-e & zscore & elbow-1 & elbow-e & zscore \\ 
  \midrule
 10 & \textbf{0.85} & \textbf{0.85} & 0.69 & 0.78 & 0.79 & 0.80 \\ 
   20 & \textbf{0.68} & \textbf{0.68} & 0.30 & 0.65 & 0.65 & 0.65 \\ 
   30 & 0.59 & 0.59 & 0.41 & \textbf{0.68} & 0.67 & 0.69 \\ 
   40 & 0.40 & 0.40 & 0.25 & 0.58 & 0.57 & \textbf{0.60} \\ 
   50 & 0.36 & 0.35 & 0.22 & \textbf{0.72} & 0.70 & 0.71 \\ 
   75 & 0.35 & 0.34 & 0.23 & 0.73 & 0.72 & \textbf{0.75} \\ 
  100 & 0.26 & 0.26 & 0.17 & 0.62 & 0.61 & \textbf{0.64} \\ 
   \bottomrule
\end{tabular}
\end{small}
\caption{Outlier explanation performance (F1 score) for synthetic data with different search strategies and dimensionality selection methods. elbow-1: elbow dimensionality selection with $\kappa=1$; elbow-e: with $\kappa=\text{exp}(1)$; zscore: z-score dimensionality selection. }
\label{tbl:dim-selection-res}
\end{table}

To assess \textbf{Q3}, we compared the z-score-based method for dimensionality selection with the elbow method, with thresholds $\kappa=1$ and $\kappa=\text{exp}(1)$. The outlier explanation results are shown in Table  \ref{tbl:dim-selection-res}.

For the elbow method, the results are insensitive to the value of $\kappa$, in both forward beam search as well as backward elimination.  This observation is consistent with the intuition given in Figure \ref{fig:steep-ll}: The difference in log likelihood between the true explanatory subspace and any lower-dimensional projection is often large, such that the actual value of $\kappa$ is less relevant.
Furthermore, results of the elbow method are not worse (and sometimes even better for forward beam search) than the z-score method, while being computationally less expensive (the z-score requires to compute outlier scores of all training samples in all investigated subspaces, while the elbow method does not).

Overall, the results show that the elbow method viable alternative to the conventional z-score transformation to select the dimensionality of the explanation.

\subsection{Runtime}

\begin{figure}
\centering
\includegraphics[scale=0.53]{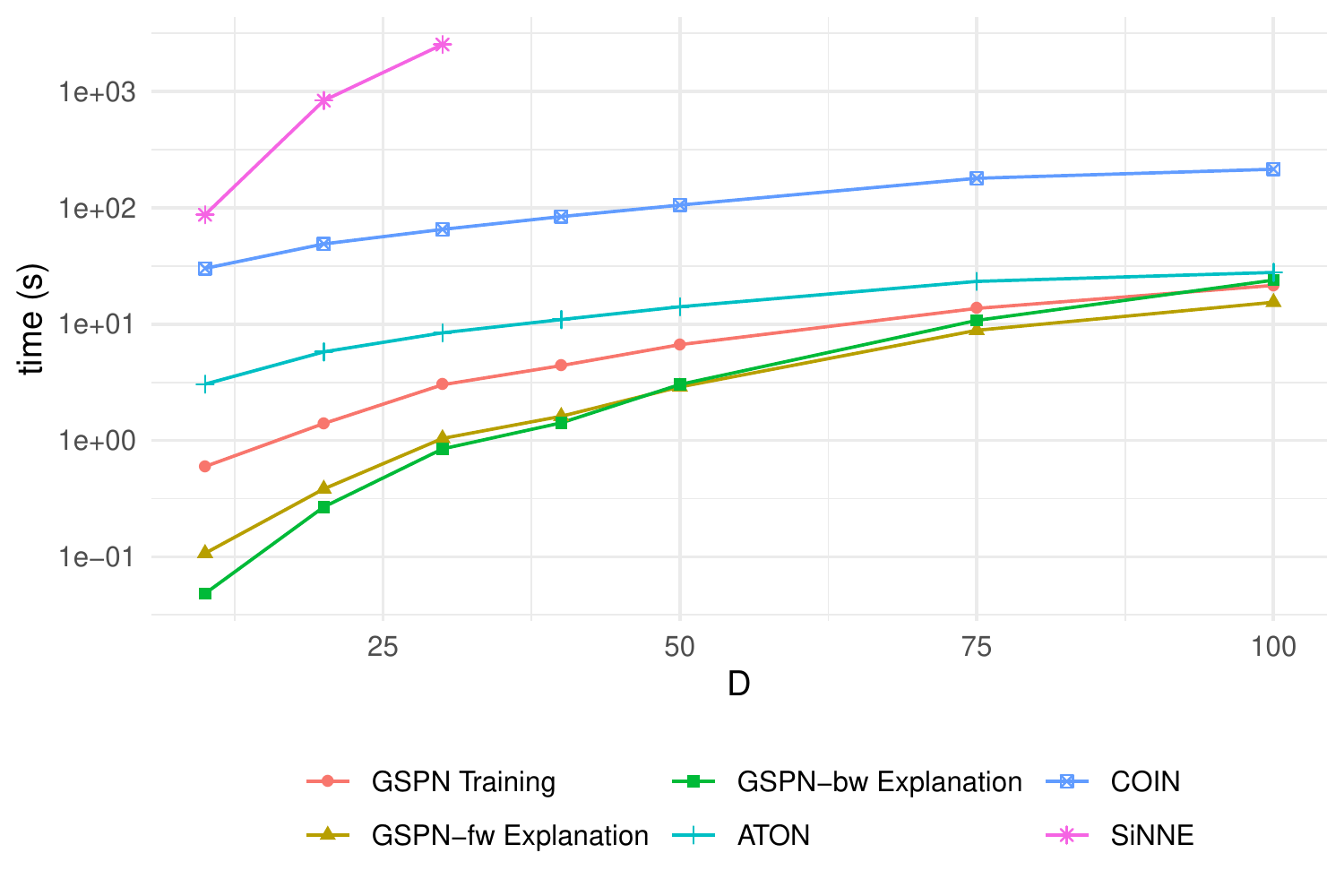}
\caption{Runtime of the different outlier explanation methods. For each dataset and method, we measured the overall runtime of explaining all outliers of that dataset.}
\label{fig:runtime}
\end{figure}

Finally, we compared the runtime of SPN-bw and SPN-fw (both using the elbow method for selecting the explanation dimensionality) with runtime of the state-of-the-art models. 
 All models were trained and evaluated on an 8-core laptop CPU (Intel Core i7-10510U). 
Figure \ref{fig:runtime} shows the runtime of the models for varying data dimensionality. 

For a moderate number of dimensions ($D \leq 100$), our SPN-based methods have a substantially lower explanation generation runtime than the other search-based method (SiNNE) and comparable runtime to the deep learning-based method (ATON).
 Note that  both ATON as well as the SPNs could also be evaluated on a GPU, which could influence relative performance. Specifically, \citet{peharz2020einsum} recently proposed an efficient GPU implementation of SPNs which is sometimes orders of magnitude faster than other implementations.

Overall, the results indicate the computational feasibility of our approach and competitiveness to state-of-the-art methods.

\section{Discussion and Conclusion}

In this paper, we proposed to use Sum-Product Networks (SPNs) for outlier detection and explanation. 
SPNs can model high-dimensional, mixed discrete-continuous distributions accurately and efficiently. Due to the tractability of marginal inference of SPNs, identifying explanations (feature subsets in which a sample is specifically anomalous) becomes efficient, allowing the used of backward elimination search. 
We empirically showed that our approach can generate more accurate explanations than existing methods, clearly outperforming other search-based approaches, and even outperforming deep learning-based methods in the majority of the cases.

Here, we only investigated outlier explanation for tabular data. Applying SPNs to the closely related task of \emph{image anomaly localization} \cite{venkataramanan2020attention} is a possible next step. 
For this task, SPNs suitable for images (like Deep Convolutional SPNs \cite{butz2019deep}) together with efficient SPN training algorithms and implementations (like the recently proposed Einsum Networks \cite{peharz2020einsum}) are an attractive option.


\bibliography{biblio}

\end{document}